\documentclass[Afour,sageh,times]{sagej}

\usepackage{xurl}
\usepackage{moreverb}
\usepackage{ulem}

\usepackage[colorlinks,bookmarksopen,bookmarksnumbered,citecolor=red,urlcolor=red]{hyperref}
\usepackage{subcaption}
\usepackage{multirow}
\usepackage{siunitx}

\newcommand\BibTeX{{\rmfamily B\kern-.05em \textsc{i\kern-.025em b}\kern-.08em
T\kern-.1667em\lower.7ex\hbox{E}\kern-.125emX}}

\def\volumeyear{2023}
\begin{document}

\runninghead{Effects of Object Features on Pick-and-Place}
\title{The Effects of Selected Object Features on a Pick-and-Place Task: a Human Multimodal Dataset}

\author{Linda Lastrico\affilnum{1,2}, Valerio Belcamino\affilnum{1}, Alessandro Carf\`i\affilnum{1}, Alessia Vignolo\affilnum{2}, Alessandra Sciutti\affilnum{2}, Fulvio Mastrogiovanni\affilnum{1}, Francesco Rea\affilnum{2}}

\affiliation{\affilnum{1} The Engine Room, Department of Informatics, Bioengineering, Robotics, and Systems Engineering (DIBRIS), University of Genoa, Italy\\
\affilnum{2} Cognitive Architecture for Collaborative Technologies Unit (CONTACT), Italian Institute of Technology, Italy}

\corrauth{Alessandro Carf\`i, Department of Informatics, Bioengineering, Robotics, and Systems Engineering (DIBRIS), University of Genoa, Genoa, Viale Causa 13, Italy}

\email{alessandro.carfi@dibris.unige.it}

\begin{abstract}
We propose a dataset to study the influence of object-specific characteristics on human pick-and-place movements and compare the quality of the motion kinematics extracted by various sensors. This dataset is also suitable for promoting a broader discussion on general learning problems in the hand-object interaction domain, such as intention recognition or motion generation with applications in the Robotics field.
The dataset consists of the recordings of 15 subjects performing 80 repetitions of a pick-and-place action under various experimental conditions, for a total of 1200 pick-and-places. The data has been collected thanks to a multimodal set-up composed of multiple cameras, observing the actions from different perspectives, a motion capture system, and a wrist-worn inertial measurement unit. All the objects manipulated in the experiments are identical in shape, size, and appearance but differ in weight and liquid filling, which influences the carefulness required for their handling. 

\end{abstract}

\maketitle

\section{Introduction}
\label{sec:introduction}

The physical characteristics of objects can play a significant role in how humans handle them. 
It has been shown that the weight of an object directly relates to the speed adopted to lift and transport it, therefore influencing the kinematics of the movement \citep{velWeight, weightFlanagan}. 
Similarly, when engaging with an item difficult to transport without damage, we approach it carefully and move it with particular attention, reducing the speed and prolonging the deceleration phase \citep{otherCareful, HFR}. 
Human-to-human communication heavily relies on implicit non-verbal signals for coordination and intention understanding. Therefore, how an object is handled conveys relevant information to the observers.

When dealing with robots, humans tend to attribute human-like abilities to them.
For example, we expect a robot to focus its attention on where its cameras are looking or be immediately ready to hand over an object we need. 
When our expectations are disappointed, the whole interaction can be compromised \citep{breazeal, chaminade, Sandini}. 
Furthermore, for robots to collaborate with humans, it is fundamental to understand what the partner is doing and when it is the right moment to act. 
In this context, it is necessary to study how humans communicate through implicit clues to enhance the interaction capabilities of robots \citep{signaling, legibility}.

Recently, there has been a growth of research interest in solutions to estimate the physical characteristics of objects when humans manipulate them.
The main insight is that a correct estimate of object properties would allow a robot to interact with a human more appropriately, especially when physical interactions are involved, e.g., a handover. 
Researchers explored the usage of deep neural networks to estimate containers capacity, dimension, and weight while observing human manipulations with RGB-D \citep{wang2022improving,apicella2022container} or even simple RGB cameras \citep{pang2021towards}. 
However, a robot should also consider if an object requires particular care for being handled to avoid changing irreparably some of its properties. 
The state-of-the-art describes this problem as a binary classification, whereby the robot should distinguish whether or not the object manipulation requires carefulness.
In this context, proposed solutions include template matching approaches relying on Gaussian Mixture Models \citep{otherCareful} or deep neural network classifiers \citep{HFR}. The definition of carefulness is not straightforward, since many factors may solicit cautious actions: from the physical context where the action takes place, to the properties of the object involved such as its fragility, precarious balance, or content about to spill \citep{Lastrico2022ICDL}. Given the difficulty in framing this feature, not many studies explicitly refer to it, although it is addressed, for example, in the handling of filled containers \citep{xompero2022corsmal,Mottaghi2017glass,pang2021towards,billard:benchmark,Yu2015fill}. From the same perspective, we define carefulness as the modulation of arm motion that minimizes liquid spilling during the manipulation of containers. Therefore, in the following, we will investigate the concept of carefulness in scenarios where humans transport containers and adapt their motion to the presence or absence of liquid inside.\\
\indent Since most of these studies rely on data-driven approaches, datasets must ground the learning process.
The literature presents different examples of datasets for the study of object manipulations \citep{survey}, most of the time focusing on the interaction between the humans and the objects in specific applications, e.g., activities of daily living \citep{dailymanip}, kitchen-related actions \citep{kitchen2009,kitchen2013,moca}, or handovers \citep{carfiHandover}.
A recently published dataset, the CORSMAL Container Manipulation by \cite{xompero2022corsmal}, collects actions such as pouring and handover initiations with containers with various shapes, materials, and content recorded with RGB-D cameras and microphones. However, the datasets currently available are more oriented to classical object recognition problems, proposing a high variability in the shapes and sizes of the objects examined, considering fewer sensors or a limited pool of participants; moreover, no one takes into consideration nor is aimed at modeling carefulness.\\
\indent Therefore, the contribution of this paper is twofold. 
First, we introduce a novel dataset describing careful and non-careful manipulations and narrow the focus on human pick-and-place actions of transparent cups, with or without water filling and balanced weights. Limiting the variability of objects allows the effect of features on human motion to be studied in detail. Data are recorded with a synchronized multisensory setup, i.e., motion capture system (MoCap), cameras, wrist-worn inertial measurement units (IMUs) and the robot point of observation, allowing for a complementary description of the scene from different perspectives. We then present an example of usage of the dataset by training Long Short-Term Memory Neural Networks, using the different sources of information provided by the dataset, to discriminate if carefulness was adopted during the manipulation and if the cup transported was light or heavy. The novelty of the approach lies in using the modulations of human kinematics to foster the inference process and understand the objects' latent features, completely overlooking their appearance. Moreover, the controlled design of the setup allows for deeply studying human strategies during the pick-and-place of objects with specific properties. By identifying the general rules to act on a set of objects, independently from their use, shape, or material, they can be used to design a robot's behavior appropriately.\\
\indent The article is organized as follows. Section \ref{sec:experimentalsetup} describes, in detail, the study design and the data acquisition process. In Section \ref{sec:datarecords} we present a description of the dataset and its organization. Section \ref{sec:codeavailability} describes the code provided to inspect and visualize the dataset. Finally, Section \ref{sec:exampleofuse} presents an example of how to use the dataset. Conclusions follow.

\section{Experimental Setup} \label{sec:experimentalsetup}
This section describes the study design and its main technical characteristics. Liguria Regional Ethical Committee approved the research protocol for this study (protocol 396REG2016 of July 25\textsuperscript{th}, 2019), and all participants provided written informed consent to publish the collected data.

\subsection{Study Design}
An object may require careful manipulation for different reasons. A glass full of water requires carefulness to avoid spilling, while a ceramic vase requires carefulness to avoid breaking it. Likewise, the different reasons inducing carefulness also influence its physical manifestation.
The careful manipulation of a glass of water would manifest in slow motions with constant orientation. Instead, carefully manipulating a ceramic vase would maximize the distance to nearby objects to avoid collisions. 

Given the limitations of previous research in the field, we decided to narrow the study limiting the notion of carefulness to the one induced by the need to move a container filled with a liquid while avoiding spills. 
In particular, the human actions recorded in the dataset consist of reaching, transportation, and departing movements involving four possible objects. 
In order to allow for simple reproduction of the experiments, we chose plastic glasses, which are easy to manipulate and of everyday use.  
The glasses, identical in shape and material, differed in their contents. 
In order to induce careful behavior in the recorded actions, we filled two of the glasses with water to the brim so that they required a high level of carefulness. 
Two different weights are considered: light (W1: 167 grams) and heavy (W2: 667 grams). 
Such values were determined by the fact that we wanted the light and heavy objects to be consistently different (500 grams) while balancing the presence or the absence of water in containers with the same volume. 
The desired weights were obtained by adding screws and coins inside the glasses, until reaching W1 or W2, balancing the presence of water. 
In this way, we defined four classes of actions, depending on the properties of the manipulated object, namely light and not careful (W1-NC), light and careful (W1-C), heavy and not careful (W2-NC), heavy and careful (W2-C). 

The sequence of performed actions is the same for every participant. It is designed to alternate the manipulation of the four categories of objects together with the direction of the movements. 

\begin{figure*}[t!]
\centering
  \begin{subfigure}[b]{0.47\textwidth}
    \includegraphics[trim={0 2.14cm 0 0},clip,width=\linewidth]{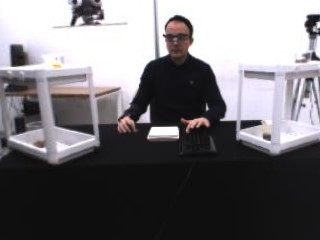}
    \caption{Robot view}
    \label{fig:restpos_robot}
  \end{subfigure}
  \hfill
  \begin{subfigure}[b]{0.47\textwidth}
    \includegraphics[width=1\linewidth]{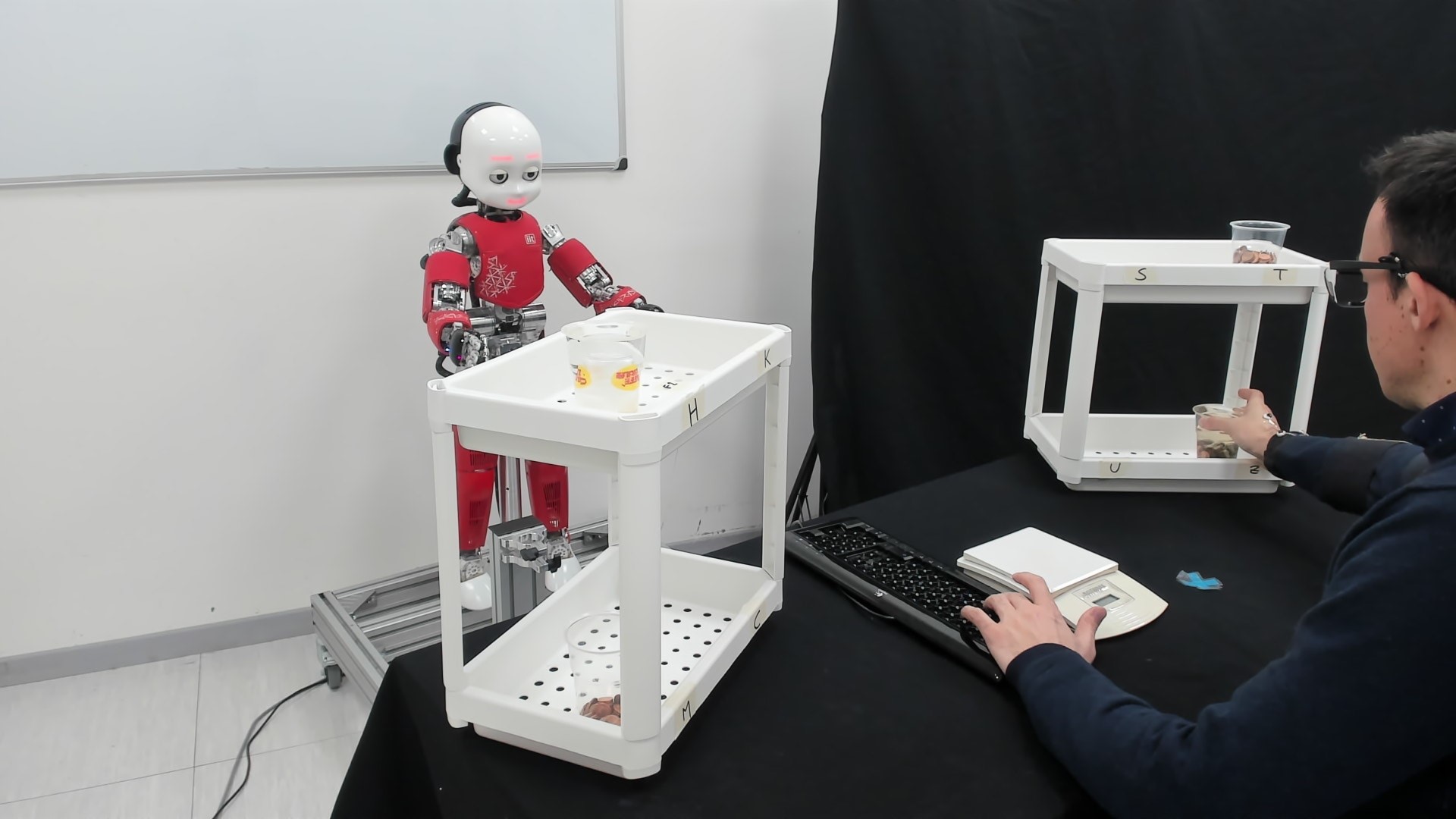}
    \caption{Lateral view}
    \label{fig:shelfpos_lat}
  \end{subfigure}
  \hfill
  \begin{subfigure}[b]{0.47\textwidth}
    \includegraphics[width=1\linewidth]{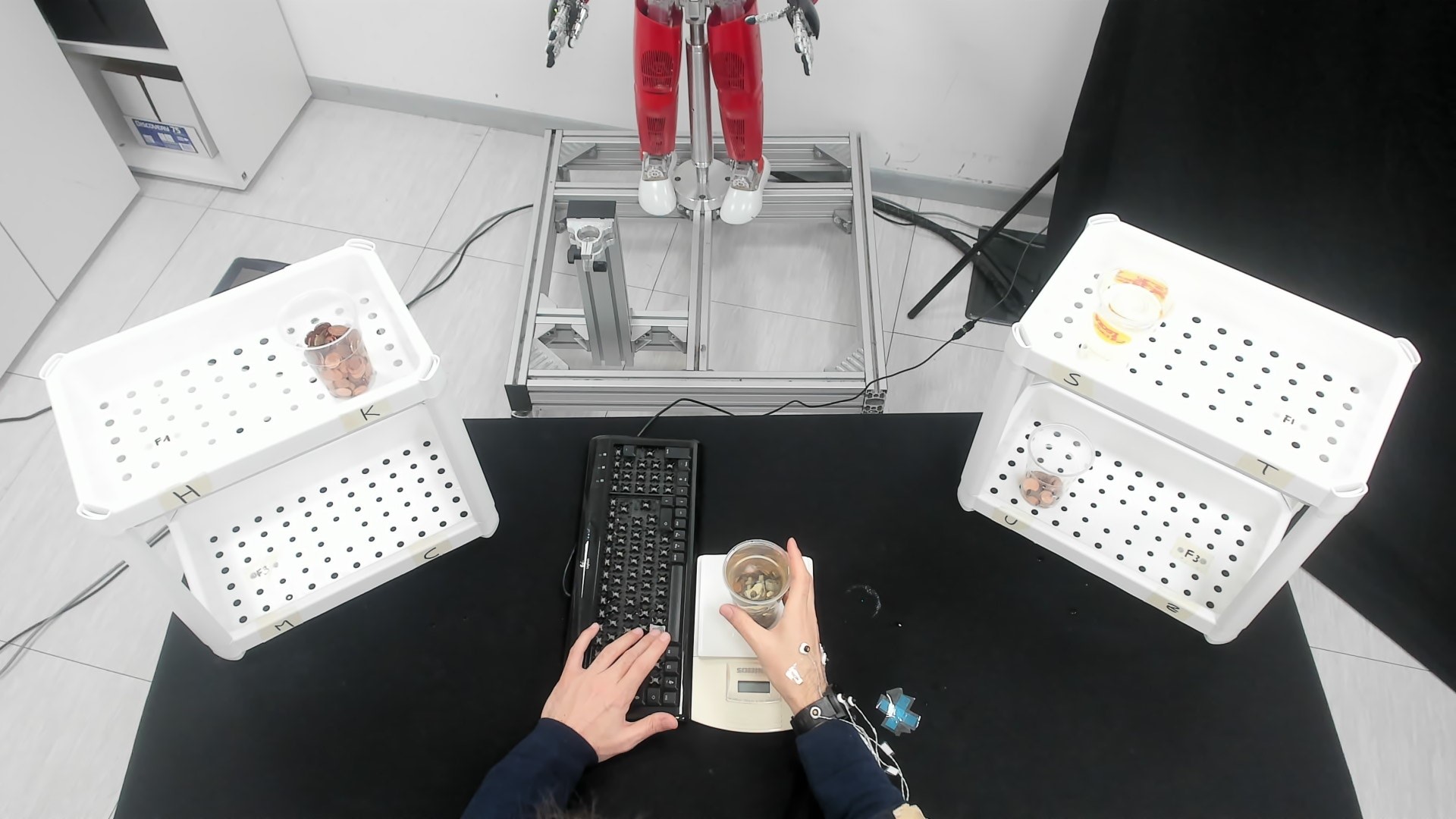}
    \caption{Back view}
    \label{fig:scalepos_back}
  \end{subfigure}
  \hfill
  \begin{subfigure}[b]{0.47\textwidth}
    \includegraphics[width=1\linewidth]{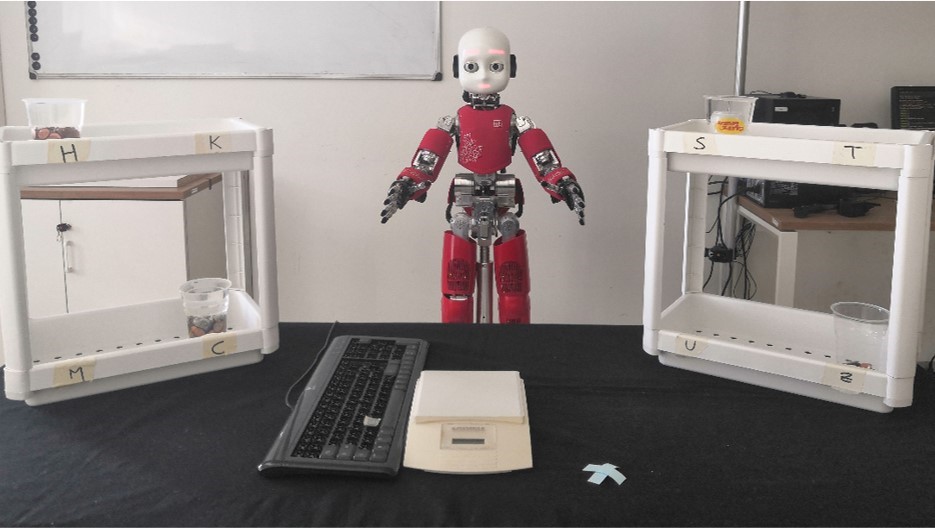}
    \caption{Frontal view}
    \label{fig:frontal}
  \end{subfigure}
  \caption{The three points of view of the experiment: the resting position as seen by the iCub robot (\ref{fig:restpos_robot}), a transportation movement towards the right shelf (\ref{fig:shelfpos_lat}), and the positioning of a glass on the scale (\ref{fig:scalepos_back}). In (\ref{fig:frontal}) the labels identify the 8 positions on the shelves.}
  \label{fig:setup}
\end{figure*}

\begin{figure}[t!]
\centering
\includegraphics[width=0.8\columnwidth]{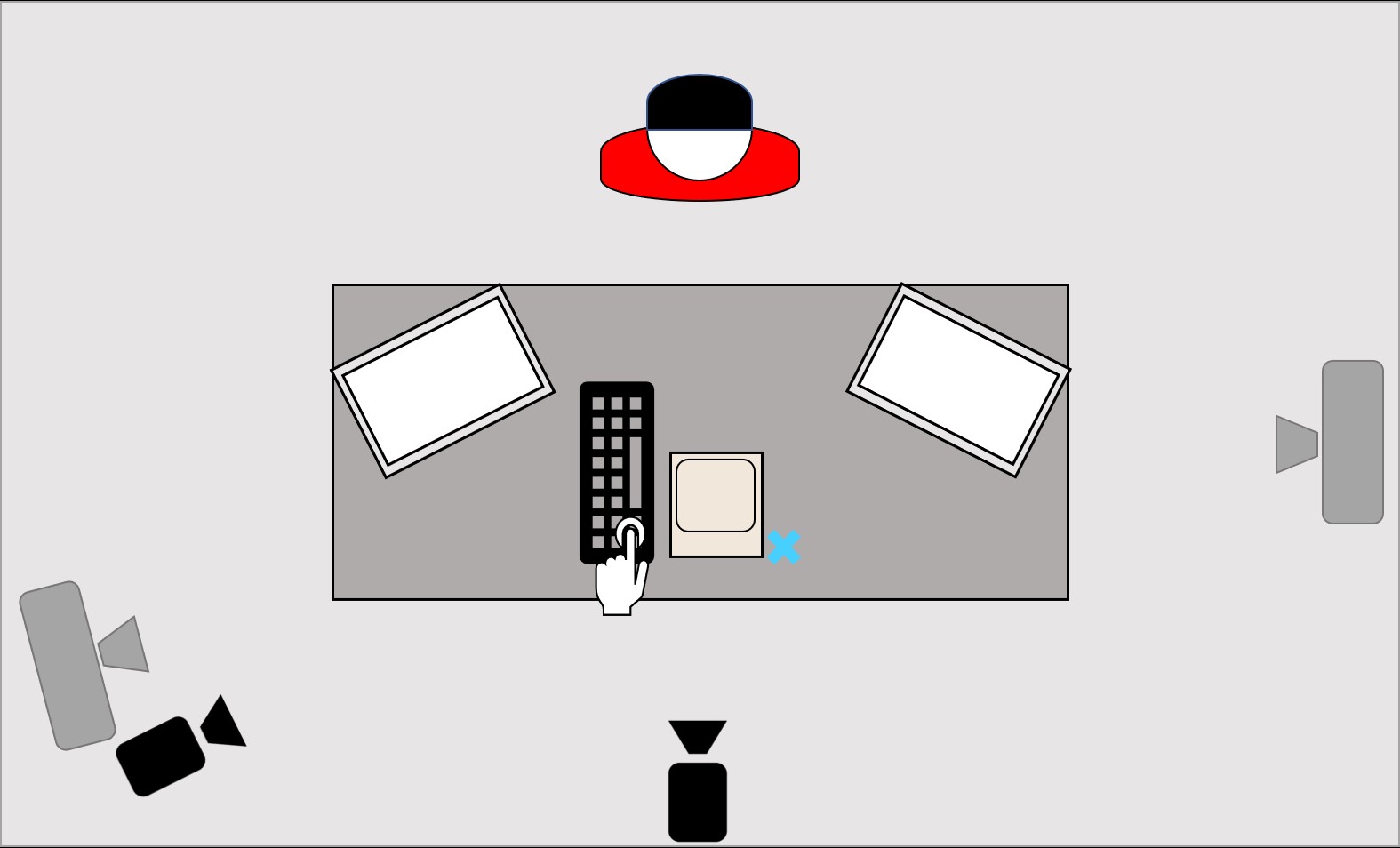}
\caption{Top view of the experimental setup. In dark grey the cameras from the Optotrak motion capture system used to detect the active markers, while in black the two high resolution cameras.}
\label{fig:schemaSetup}
\end{figure}

At the beginning of the experiment, the volunteer sits at a table, with their hands resting on it. 
On the table, covered with a black cloth, there is one shelf at each end, a scale right in front of the subject, and a keyboard on the left side, see Figure \ref{fig:setup} and Figure \ref{fig:schemaSetup}. 
Each shelf has four possible positions, denoted by a letter marked on the frontal edge of the shelf, where the glasses could be positioned, i.e., two on the bottom level and two on the upper one, see Figure \ref{fig:frontal}. 
The shelves measure $36\times23$ cm, the top level is 36 cm above the bottom one and there is a border of 6 cm, delimiting each shelf and constituting an obstacle. 
The two positions on each shelf level are indicatively 18 cm apart.
A blue cross on the table marks the resting position that the participants' right hand should reach after each movement, see Figure \ref{fig:restpos_robot}. 
The distance between the starting position and the shelves is indicatively 50 cm. 
The humanoid robot iCub \citep{metta2008icub} is placed in front of the table and passively records the scene with its left camera.

\begin{figure*}[t]
\centering
    \begin{subfigure}[b]{.9\textwidth}
        \includegraphics[width=1\linewidth]{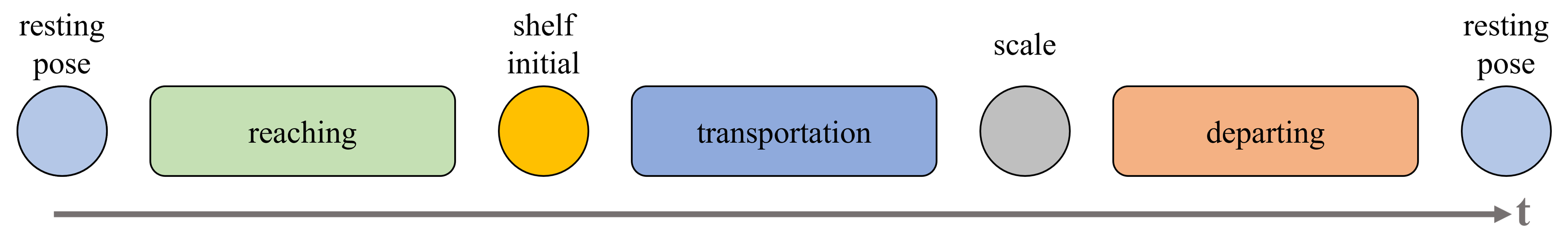}
        \subcaption{Transport from shelf to scale}
        \label{fig:fromshelf}
  \end{subfigure}
    \begin{subfigure}[b]{.9\textwidth}
        \includegraphics[width=1\linewidth]{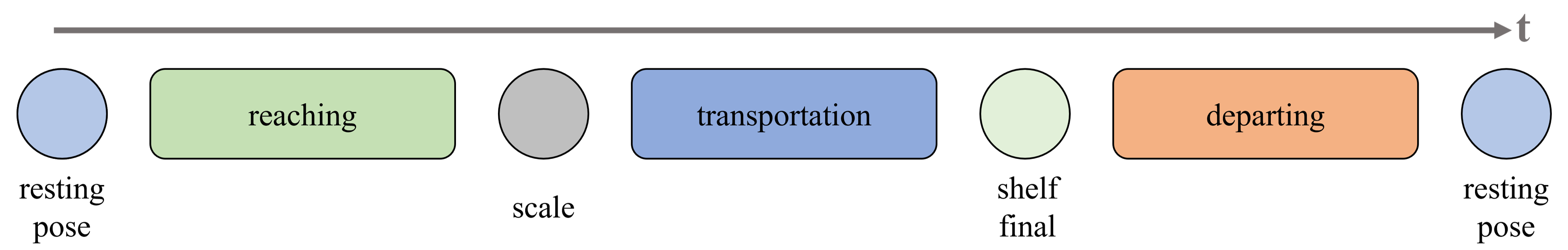}
        \subcaption{Transport from scale to shelf}
        \label{fig:fromscale}
  \end{subfigure}
\caption{The structure of each trial falls within two possibilities, whereby the main action is the glass transportation: in \ref{fig:fromshelf} are shown the steps for taking a glass from the shelves and placing it on the scale, while in \ref{fig:fromscale} those for putting back the glass from the scale to the shelf. }
\label{fig:sequence}
\end{figure*}

As instructed by a synthetic voice, the participants perform a series of reaching, transportation, and departing movements of the four glasses. 
The volunteers interacted with the items with their right hand and received instructions on the next movement to perform by pressing a key on the keyboard with their left hand.
The experiment is set up as summarized in Figure \ref{fig:sequence}:
\begin{enumerate}
    \item The experiment starts with the volunteer in the resting position and with the four objects distributed on the shelves.
    \item When a key of the keyboard is pressed, a synthetic voice indicates the position on the shelf of the object to be transported. The position is referred to using the corresponding letter.
    \item The volunteer reaches for the glass and grasps it in the specified location (reaching phase), as in Figure \ref{fig:shelfpos_lat}.
    \item In the transportation phase, the volunteer moves the glass from the shelf (shelf initial position) to the scale.
    \item The volunteer releases the glass and returns with the dominant hand to the resting position.
    \item The volunteer presses the key a second time, and the synthetic voice suggests where the glass should be placed on the other shelf. The shelf spot chosen this time is vacant.
    \item The volunteer reaches for the scale and takes the glass, see Figure \ref{fig:scalepos_back}.
    \item The volunteer moves the glass from the scale to its final location on the shelf, performing a transportation action.
    \item The volunteer places the glass down and returns to the resting position.
\end{enumerate}

The order in which the volunteers performed the experiment is detailed in Table \ref{tab:trials}. 
The first sixteen trials were used as a practice before the main experiment started. 
In the main experiment, each volunteer performed 64 reaching movements, 64 transportation movements, and 64 departing movements to the resting position.
After the 16\textsuperscript{th} and the 48\textsuperscript{th} trial, the objects' position is changed by an experimenter to maintain the properties of the manipulated objects and the initial position of the glasses equally balanced. To prevent fatigue and gesture automatism, at the end of each sequence of 16 pick-and-place actions, participants could rest as much as they wanted, both physically and mentally.

\begin{table}[t!]
\caption[Movements sequence performed by each participant (table 1/2).]{Movements sequence performed by each participant. The first column refers to the ID used to identify each trial. The second and third columns specify the characteristics of the object handled in the trial, which could be light (\textit{W1}) or heavy (\textit{W2}) and require carefulness (\textit{c}) or not (\textit{NC}). The last two columns refer to object positions: they indicate whether the glass to grab was placed on the shelf or on the scale, and the shelf position where the glass needed to be taken from or put down.}
\label{tab:trials}
\begin{center}
\resizebox{0.9\columnwidth}{!}{\begin{tabular}{cccccc}
\toprule
\multicolumn{1}{c}{} & \multicolumn{1}{c}{\textbf{Trial ID}} & \multicolumn{1}{c}{\textbf{Weight}} & \multicolumn{1}{c}{\textbf{Care}} & \multicolumn{1}{c}{\begin{tabular}[c]{@{}c@{}}\textbf{Object Initial}\\ \textbf{Position}\end{tabular}} & \multicolumn{1}{c}{\begin{tabular}[c]{@{}c@{}}\textbf{Shelf}\\ \textbf{Position}\end{tabular}} \\ 
\toprule
\multicolumn{1}{c}{\multirow{16}{*}{\rotatebox{90}{
\textit{Practice}}}} & \multicolumn{1}{c}{1}  & \multicolumn{1}{c}{W1} & \multicolumn{1}{c}{NC} & \multicolumn{1}{c}{Shelf}  & \multicolumn{1}{c}{Z}\\ 
\multicolumn{1}{c}{} & \multicolumn{1}{c}{2}  & \multicolumn{1}{c}{W1} & \multicolumn{1}{c}{NC} & \multicolumn{1}{c}{Scale}  & \multicolumn{1}{c}{K}\\
\multicolumn{1}{c}{} & \multicolumn{1}{c}{3}  & \multicolumn{1}{c}{W2} & \multicolumn{1}{c}{NC} & \multicolumn{1}{c}{Shelf}  & \multicolumn{1}{c}{H}\\
\multicolumn{1}{c}{} & \multicolumn{1}{c}{4}  & \multicolumn{1}{c}{W2} & \multicolumn{1}{c}{NC} & \multicolumn{1}{c}{Scale}  & \multicolumn{1}{c}{U}\\
\multicolumn{1}{c}{} & \multicolumn{1}{c}{5}  & \multicolumn{1}{c}{W1} & \multicolumn{1}{c}{C} & \multicolumn{1}{c}{Shelf}  & \multicolumn{1}{c}{S}\\
\multicolumn{1}{c}{} & \multicolumn{1}{c}{6}  & \multicolumn{1}{c}{W1} & \multicolumn{1}{c}{C} & \multicolumn{1}{c}{Scale}  & \multicolumn{1}{c}{M}\\
\multicolumn{1}{c}{} & \multicolumn{1}{c}{7}  & \multicolumn{1}{c}{W2} & \multicolumn{1}{c}{C} & \multicolumn{1}{c}{Shelf}  & \multicolumn{1}{c}{C}\\
\multicolumn{1}{c}{} & \multicolumn{1}{c}{8}  & \multicolumn{1}{c}{W2} & \multicolumn{1}{c}{C} & \multicolumn{1}{c}{Scale}  & \multicolumn{1}{c}{T}\\
\multicolumn{1}{c}{} & \multicolumn{1}{c}{9}  & \multicolumn{1}{c}{W1} & \multicolumn{1}{c}{NC} & \multicolumn{1}{c}{Shelf}  & \multicolumn{1}{c}{K}\\
\multicolumn{1}{c}{} & \multicolumn{1}{c}{10}  & \multicolumn{1}{c}{W1} & \multicolumn{1}{c}{NC} & \multicolumn{1}{c}{Scale}  & \multicolumn{1}{c}{Z}\\
\multicolumn{1}{c}{} & \multicolumn{1}{c}{11}  & \multicolumn{1}{c}{W2} & \multicolumn{1}{c}{NC} & \multicolumn{1}{c}{Shelf}  & \multicolumn{1}{c}{U}\\
\multicolumn{1}{c}{} & \multicolumn{1}{c}{12}  & \multicolumn{1}{c}{W2} & \multicolumn{1}{c}{NC} & \multicolumn{1}{c}{Scale}  & \multicolumn{1}{c}{H}\\
\multicolumn{1}{c}{} & \multicolumn{1}{c}{13}  & \multicolumn{1}{c}{W1} & \multicolumn{1}{c}{C} & \multicolumn{1}{c}{Shelf}  & \multicolumn{1}{c}{M}\\
\multicolumn{1}{c}{} & \multicolumn{1}{c}{14}  & \multicolumn{1}{c}{W1} & \multicolumn{1}{c}{C} & \multicolumn{1}{c}{Scale}  & \multicolumn{1}{c}{S}\\
\multicolumn{1}{c}{} & \multicolumn{1}{c}{15}  & \multicolumn{1}{c}{W2} & \multicolumn{1}{c}{C} & \multicolumn{1}{c}{Shelf}  & \multicolumn{1}{c}{T}\\
\multicolumn{1}{c}{} &\multicolumn{1}{c}{16}  & \multicolumn{1}{c}{W2} & \multicolumn{1}{c}{C} & \multicolumn{1}{c}{Scale}  & \multicolumn{1}{c}{C}\\ 
\midrule
\multicolumn{1}{c}
{\multirow{16}{*}{\rotatebox{90}{\textit{1\textsuperscript{st} Session}}}} & \multicolumn{1}{c}{17, 33}  & \multicolumn{1}{c}{W1} & \multicolumn{1}{c}{NC} & \multicolumn{1}{c}{Shelf}  & \multicolumn{1}{c}{H}\\ 
\multicolumn{1}{c}{} & \multicolumn{1}{c}{18, 34}  & \multicolumn{1}{c}{W1} & \multicolumn{1}{c}{NC} & \multicolumn{1}{c}{Scale}  & \multicolumn{1}{c}{U}\\
\multicolumn{1}{c}{} & \multicolumn{1}{c}{19, 35}  & \multicolumn{1}{c}{W2} & \multicolumn{1}{c}{NC} & \multicolumn{1}{c}{Shelf}  & \multicolumn{1}{c}{Z}\\
\multicolumn{1}{c}{} & \multicolumn{1}{c}{20, 36}  & \multicolumn{1}{c}{W2} & \multicolumn{1}{c}{NC} & \multicolumn{1}{c}{Scale}  & \multicolumn{1}{c}{K}\\
\multicolumn{1}{c}{} & \multicolumn{1}{c}{21, 37}  & \multicolumn{1}{c}{W1} & \multicolumn{1}{c}{C} & \multicolumn{1}{c}{Shelf}  & \multicolumn{1}{c}{M}\\
\multicolumn{1}{c}{} & \multicolumn{1}{c}{22, 38}  & \multicolumn{1}{c}{W1} & \multicolumn{1}{c}{C} & \multicolumn{1}{c}{Scale}  & \multicolumn{1}{c}{S}\\
\multicolumn{1}{c}{} & \multicolumn{1}{c}{23, 39}  & \multicolumn{1}{c}{W2} & \multicolumn{1}{c}{C} & \multicolumn{1}{c}{Shelf}  & \multicolumn{1}{c}{T}\\
\multicolumn{1}{c}{} & \multicolumn{1}{c}{24, 40}  & \multicolumn{1}{c}{W2} & \multicolumn{1}{c}{C} & \multicolumn{1}{c}{Scale}  & \multicolumn{1}{c}{C}\\
\multicolumn{1}{c}{} & \multicolumn{1}{c}{25, 41}  & \multicolumn{1}{c}{W1} & \multicolumn{1}{c}{NC} & \multicolumn{1}{c}{Shelf}  & \multicolumn{1}{c}{U}\\
\multicolumn{1}{c}{} & \multicolumn{1}{c}{26, 42}  & \multicolumn{1}{c}{W1} & \multicolumn{1}{c}{NC} & \multicolumn{1}{c}{Scale}  & \multicolumn{1}{c}{H}\\
\multicolumn{1}{c}{} & \multicolumn{1}{c}{27, 43}  & \multicolumn{1}{c}{W2} & \multicolumn{1}{c}{NC} & \multicolumn{1}{c}{Shelf}  & \multicolumn{1}{c}{K}\\
\multicolumn{1}{c}{} & \multicolumn{1}{c}{28, 44}  & \multicolumn{1}{c}{W2} & \multicolumn{1}{c}{NC} & \multicolumn{1}{c}{Scale}  & \multicolumn{1}{c}{Z}\\
\multicolumn{1}{c}{} & \multicolumn{1}{c}{29, 45}  & \multicolumn{1}{c}{W1} & \multicolumn{1}{c}{C} & \multicolumn{1}{c}{Shelf}  & \multicolumn{1}{c}{S}\\
\multicolumn{1}{c}{} & \multicolumn{1}{c}{30, 46}  & \multicolumn{1}{c}{W1} & \multicolumn{1}{c}{C} & \multicolumn{1}{c}{Scale}  & \multicolumn{1}{c}{M}\\
\multicolumn{1}{c}{} & \multicolumn{1}{c}{31, 47}  & \multicolumn{1}{c}{W2} & \multicolumn{1}{c}{C} & \multicolumn{1}{c}{Shelf}  & \multicolumn{1}{c}{C}\\
\multicolumn{1}{c}{} & \multicolumn{1}{c}{32, 48}  & \multicolumn{1}{c}{W2} & \multicolumn{1}{c}{C} & \multicolumn{1}{c}{Scale}  & \multicolumn{1}{c}{T}\\
\midrule
\multicolumn{1}{c}{\multirow{16}{*}{\rotatebox{90}{\textit{2\textsuperscript{nd} Session}}}} & \multicolumn{1}{c}{49, 65}  & \multicolumn{1}{c}{W1} & \multicolumn{1}{c}{NC} & \multicolumn{1}{c}{Shelf}  & \multicolumn{1}{c}{S}\\ 
\multicolumn{1}{c}{} & \multicolumn{1}{c}{50, 66}  & \multicolumn{1}{c}{W1} & \multicolumn{1}{c}{NC} & \multicolumn{1}{c}{Scale}  & \multicolumn{1}{c}{M}\\
\multicolumn{1}{c}{} & \multicolumn{1}{c}{51, 67}  & \multicolumn{1}{c}{W2} & \multicolumn{1}{c}{NC} & \multicolumn{1}{c}{Shelf}  & \multicolumn{1}{c}{C}\\
\multicolumn{1}{c}{} & \multicolumn{1}{c}{52, 68}  & \multicolumn{1}{c}{W2} & \multicolumn{1}{c}{NC} & \multicolumn{1}{c}{Scale}  & \multicolumn{1}{c}{T}\\
\multicolumn{1}{c}{} & \multicolumn{1}{c}{53, 69}  & \multicolumn{1}{c}{W1} & \multicolumn{1}{c}{C} & \multicolumn{1}{c}{Shelf}  & \multicolumn{1}{c}{U}\\
\multicolumn{1}{c}{} & \multicolumn{1}{c}{54, 70}  & \multicolumn{1}{c}{W1} & \multicolumn{1}{c}{C} & \multicolumn{1}{c}{Scale}  & \multicolumn{1}{c}{H}\\
\multicolumn{1}{c}{} & \multicolumn{1}{c}{55, 71}  & \multicolumn{1}{c}{W2} & \multicolumn{1}{c}{C} & \multicolumn{1}{c}{Shelf}  & \multicolumn{1}{c}{K}\\
\multicolumn{1}{c}{} & \multicolumn{1}{c}{56, 72}  & \multicolumn{1}{c}{W2} & \multicolumn{1}{c}{C} & \multicolumn{1}{c}{Scale}  & \multicolumn{1}{c}{Z}\\
\multicolumn{1}{c}{} & \multicolumn{1}{c}{57, 73}  & \multicolumn{1}{c}{W1} & \multicolumn{1}{c}{NC} & \multicolumn{1}{c}{Shelf}  & \multicolumn{1}{c}{M}\\
\multicolumn{1}{c}{} & \multicolumn{1}{c}{58, 74}  & \multicolumn{1}{c}{W1} & \multicolumn{1}{c}{NC} & \multicolumn{1}{c}{Scale}  & \multicolumn{1}{c}{S}\\
\multicolumn{1}{c}{} & \multicolumn{1}{c}{59, 75}  & \multicolumn{1}{c}{W2} & \multicolumn{1}{c}{NC} & \multicolumn{1}{c}{Shelf}  & \multicolumn{1}{c}{T}\\
\multicolumn{1}{c}{} & \multicolumn{1}{c}{60, 76}  & \multicolumn{1}{c}{W2} & \multicolumn{1}{c}{NC} & \multicolumn{1}{c}{Scale}  & \multicolumn{1}{c}{C}\\
\multicolumn{1}{c}{} & \multicolumn{1}{c}{61, 77}  & \multicolumn{1}{c}{W1} & \multicolumn{1}{c}{C} & \multicolumn{1}{c}{Shelf}  & \multicolumn{1}{c}{H}\\
\multicolumn{1}{c}{} & \multicolumn{1}{c}{62, 78}  & \multicolumn{1}{c}{W1} & \multicolumn{1}{c}{C} & \multicolumn{1}{c}{Scale}  & \multicolumn{1}{c}{U}\\
\multicolumn{1}{c}{} & \multicolumn{1}{c}{63, 79}  & \multicolumn{1}{c}{W2} & \multicolumn{1}{c}{C} & \multicolumn{1}{c}{Shelf}  & \multicolumn{1}{c}{Z}\\
\multicolumn{1}{c}{} & \multicolumn{1}{c}{64, 80}  & \multicolumn{1}{c}{W2} & \multicolumn{1}{c}{C} & \multicolumn{1}{c}{Scale}  & \multicolumn{1}{c}{K}
\end{tabular}}\\
\end{center}
\end{table}

\subsection{Data Acquisition}
The data collection process involved 15 healthy right-handed participants (8 males, 7 females, age: $28.6\pm 3.9$). The participants are part of our research organizations, but none of them is directly involved in this research. Each subject performed 80 trials, ensuring 300 interactions for each of the 4 objects. Such participants' numerosity is generally higher than that of other object manipulation datasets in the literature, as highlighted by the review of \cite{survey} and, as recent examples, the works by \cite{Kratzer2021Mogaze, xompero2022corsmal} and \cite{fan2023arctic}.

The framework adopted during the experiment was designed to ensure a high degree of automation in the acquisition phase. 
Most sensors were directly interfaced with the YARP middleware \citep{yarp}, allowing timestamp synchronization, whereas the wrist-worn IMU was using a Robot Operating System (ROS)-YARP interface.

The data have been segmented to separate each action presented in Figure \ref{fig:sequence}. The MoCap and the IMU segmentation have been performed automatically, exploiting the participants' pressures on the key. 
Instead, the camera images were organized offline into separate folders according to the saved timestamps.
For each participant, we saved a log file containing information on the experiment. These log files contain the YARP timestamp of each instruction communicated by the synthetic voice and the time for each key's pressures.

\begin{figure}
    \centering
    \includegraphics[width=\columnwidth]{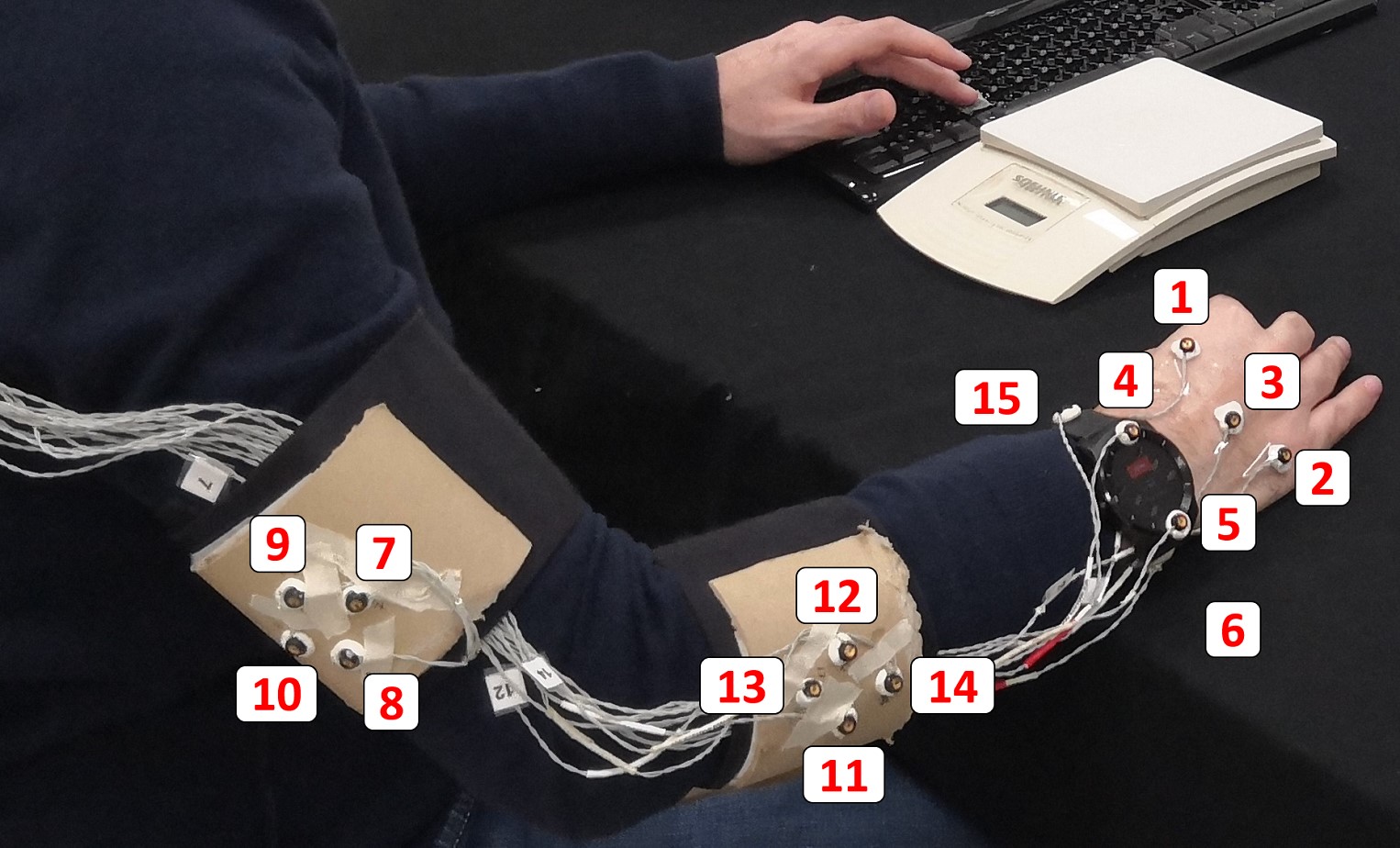}
    \caption{Marker positions on the participant's right arm and hand. The smartwatch equipped with inertial sensors is visible on the wrist.}
    \label{fig:markers}
\end{figure}

\subsubsection{Motion Capture System}
As a motion capture system (MoCap), we used the Optotrak Certus\textsuperscript{\textregistered}, NDI, with active infrared markers. 
In total, we recorded the signal from 15 markers.
As pictured in Figure \ref{fig:markers}, the five markers on the hand were placed, respectively, on the metacarpophalangeal joints of the index and the little finger, on the diaphysis of the third metacarpal, and on the smartwatch in correspondence to the radial and ulnar styloid.
Additionally, two markers were positioned on the watch strap, one per side, to better characterize wrist movements. 
Even though the main focus of the recording was to acquire hand and wrist motions, we decided to position a few markers on the participants' arm and forearm. 
We used two rigid cardboards where we put four markers each. See Figure \ref{fig:markers} for reference. 
The frequency of the acquisition is 100 $Hz$. 
For every frame, the three-dimensional coordinates of every marker (in millimeters) were saved into the file associated with the trial. 
Moreover, the timestamp at the beginning and the end of each trial was saved.
This was used to retrieve the timestamp corresponding to each frame by applying a linear interpolation.

\subsubsection{Inertial Sensors}
On the right wrist of the volunteers, we mounted an LG G Watch R smartwatch equipped with a 6-axis IMU. 
The sampling rate was 71 $Hz$. 
As for the MoCap data, a separate file was created for each trial, whenever the key on the keyboard was pressed at the end of the departing movement.
The file was saved in \texttt{json} format containing the ROS and YARP timestamps at each sample, the internal Android timestamp, and the three components of the linear acceleration in $m/s\textsuperscript{2}$ and those of the angular velocity in $rad/s$. 
The data were published by the Android app on ROS for the smartwatch acquisition and then saved. 
Through the ROS-YARP interface, the corresponding YARP timestamp was sent to ROS at each key's pressure and written on the \texttt{json} file. 

\subsubsection{Cameras}
Two cameras with a resolution of $1920\times1080$ pixels were positioned in the room where the experiments took place. 
The former was placed at the back of the participant's chair and recorded the scene from an overhead viewpoint, remaining elevated by 130 $cm$ with respect to the table. 
The latter was on the left side of the participants, with an oblique point of view, 65 $cm$ higher than the table with a distance from the hand starting position of \textit{circa} 140 $cm$. 
The reader is referred to the scheme in Figure \ref{fig:schemaSetup} for reference.
The frame rate was set to 30 $Hz$.
The last sensor used in the experiment was iCub's left camera.
The robot was located opposite the table, in front of the volunteer, with a complete perspective of the table and the shelves.
The camera's resolution was $320\times240$ pixels, and the frame rate was 22 $Hz$. Even though such resolution is particularly low, this third camera offers a complementary point of view to the other two, which is relevant for a possible deployment of the data acquired in a human-robot interaction context. However, we suggest future researchers interested in acquiring a similar setup, to use a standard webcam for the frontal view in addition to the robot one, to grant the best quality view from all perspectives.

As previously mentioned, the images acquired with the cameras were not automatically segmented during the acquisition.
Segmentation occurred afterward using the YARP timestamps, which were saved for each image from each camera in a log file.
Indeed, each saved camera frame was associated with a YARP timestamp, making it possible to relate the acquired images with the events triggered by the key's pressure.

\section{Data Records}\label{sec:datarecords}
The dataset is available on Kaggle\footnote{\raggedright \url{www.kaggle.com/dataset/cec218d6597e7c2cac28c7d6a1e8cbd381e451a77192c16b648d2b4c5de70697}}, while the software utilities can be found on GitHub\footnote{\url{https://github.com/lindalastrico/objectsManipulationDataset}}. 
While we used MATLAB as a reference software to create the utilities functions, we chose to make sensory data available in a non-proprietary format (\texttt{log}, \texttt{csv}, \texttt{json}, or \texttt{jpg}, depending on the sensor). 
A comprehensive table with a summary of performed movements, with the details about their direction and the properties of the object involved is available in the utilities, also with a machine readable structure. 
Table \ref{tab:trials} summarizes the characteristics of each trial.
The same sequence was performed by each participant. 

In the repository, the data are organized into separate folders on a sensor basis.
Inertial and MoCap recordings can be found in folders \texttt{data/inertial} and \texttt{data/mocap}, respectively. While cameras recording are divided in three folders according to the following naming:

\begin{itemize}
    \item \texttt{data/cam\textunderscore0}: low resolution frontal images from the iCub left camera,
    \item \texttt{data/cam\textunderscore1}: high resolution lateral camera,
    \item \texttt{data/cam\textunderscore2}: high resolution images from behind.
\end{itemize}

The main folder also includes one folder \texttt{data/log}, which contains the experiments log files. They report the YARP timestamp corresponding to the key's pressures at the end of each trial, together with the information relative to the transport movement: the instruction given by the synthetic voice is reported as ``sending speech: \textit{Prendi/Metti} in \textit{\{Position letter\}}''. 
The verbs ``Prendi'' and ``Metti'' mean, respectively, ``Take from'' and ``Put in'' in Italian, and were used to tell participants where to take the glass from or where to put it back on the shelf, using the letter corresponding to the position, as in Figure \ref{fig:frontal}. \\
Each one of the described folders is organized into sub-folders named from \texttt{P001} to \texttt{P015}, containing the files associated to each participant. In the case of MoCap and inertial sensors, the files are sequentially named from trial 1 to 80. \\
Instead, the camera folders are divided, for each subject, into 80 subfolders (one for each trial), e.g., the folder \path{data/cam_1/P001/P001_Trial_001} contains the images from the lateral high resolution camera for the first trial of the first subject. The same sub-folders also contain a \textit{data.txt} file reporting the correspondence between the YARP timestamps and images names for the specific trial.

The MoCap raw data are organized in \texttt{csv} files, with 62 columns and as many rows as the samples in the trial. 
The first column contains a progressive ID, the second one the YARP timestamp for each sample. 
As previously mentioned, this timestamp was computed after the acquisition by linearly interpolating the timestamps indicating the start and the end of the trial. 
The remaining columns represent triplets of three-dimensional trajectories ($x,y,z$ coordinates) followed by an additional column for each marker containing 0, if the marker was visible in that particular frame, or 1 if not. 
The order in which the markers appear in the files follows the numbering introduced in Figure \ref{fig:markers}, therefore the first triplet being the coordinates of the marker on the metacarpophalangeal joint of the index, the second triplet the marker on the joint of the little finger, and so on.

Regarding the inertial sensor raw files, they are saved in \texttt{json} format.
For each sample, the available information is the ROS timestamp, the YARP timestamp, the Android one, and the three components of linear acceleration and angular velocity.

\subsection{Data Assessment}
\begin{figure}[t]
\centering
\includegraphics[width=\columnwidth]{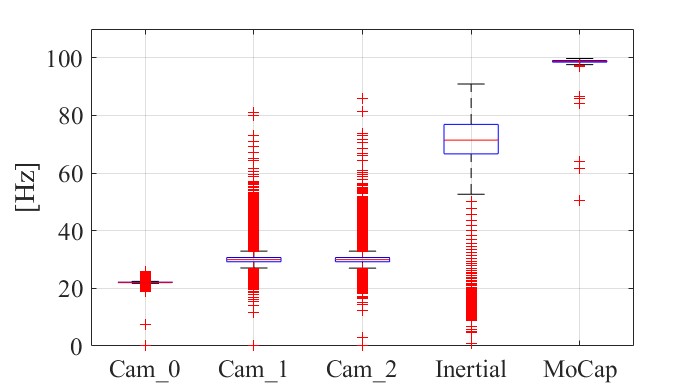}
\caption{Box plots of each sensor recording frequency. The red lines represent the medians, the blue rectangles limit the 25\textsuperscript{th} and 75\textsuperscript{th} percentiles}
\label{fig:frequencies}
\end{figure}

A first evaluation on the quality of the provided data is related to their frequency.
As already mentioned, the acquisition frequency depends on the specific sensor, and it is 100 Hz for the MoCap, 71 Hz for the inertial sensor, 30 Hz for the cameras, and 22 Hz for the robot camera. 
Figure \ref{fig:frequencies} represents the frequencies for the sensors considering the whole experiment for all the participants. Even though some outliers are present, downsampling the data to the lowest frame rate (22 Hz) allows for automatically cleaning them and keeping a frequency well above the range of human motion, that is generally below 10 Hz \citep{frequency10} and lies in the interval [$0.3, 4.5$] Hz for hand motion \citep{frequencyHand}.\\
\begin{figure}[t]
\centering
  \begin{subfigure}[b]{0.45\textwidth}
    \centering
    \includegraphics[width=\linewidth]{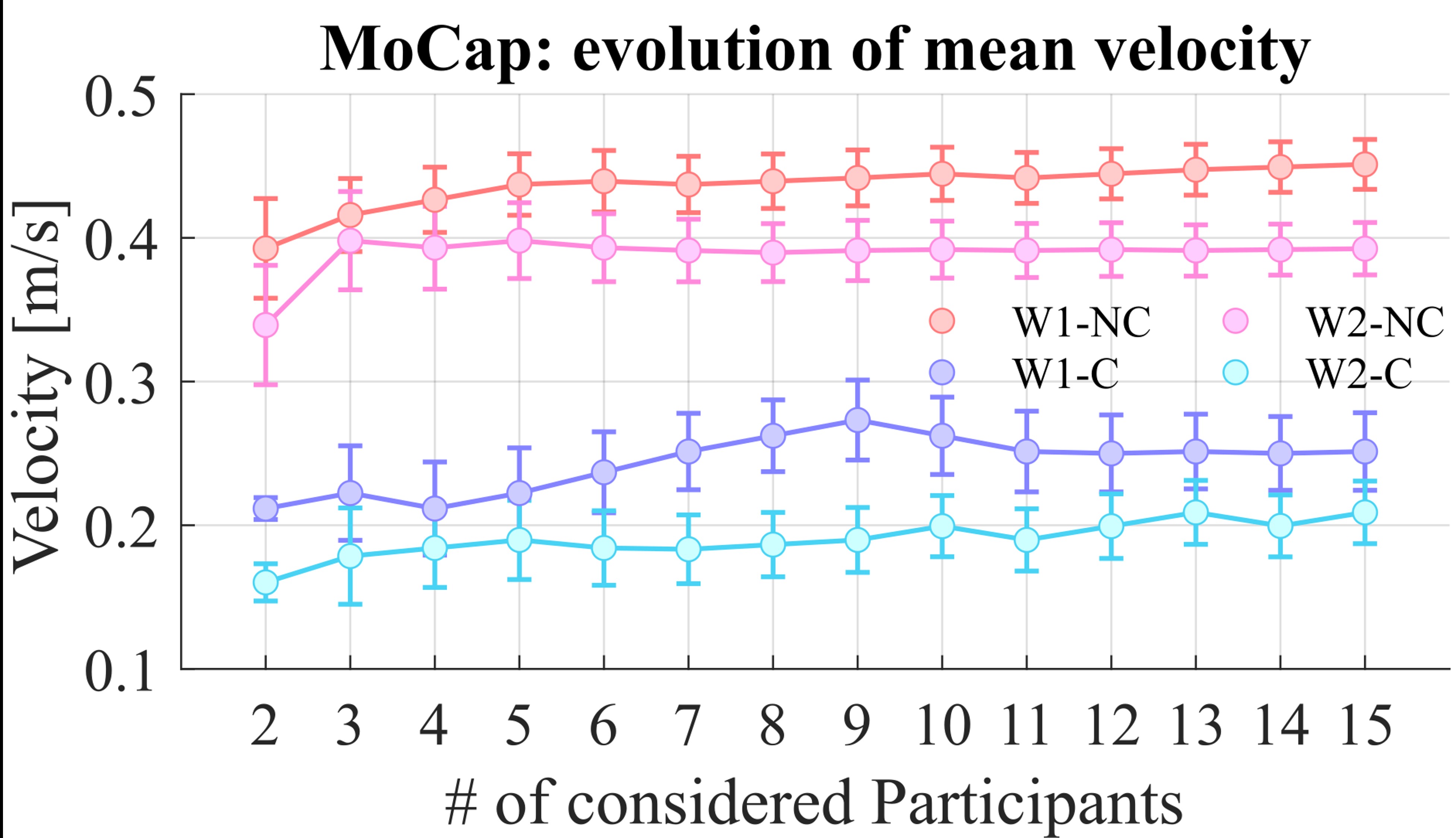}
    \caption{Mean velocity during the transport motion considering an increasing subset of participants. When considering the whole dataset, the values are the same as in Figure \ref{fig:meanVel} below. The colored dots represent the median, while the bars are the standard deviation.}
    \label{fig:evolution}
  \end{subfigure}
  \hfill
  \begin{subfigure}[b]{0.45\textwidth}
    \centering
    \includegraphics[width=\linewidth]{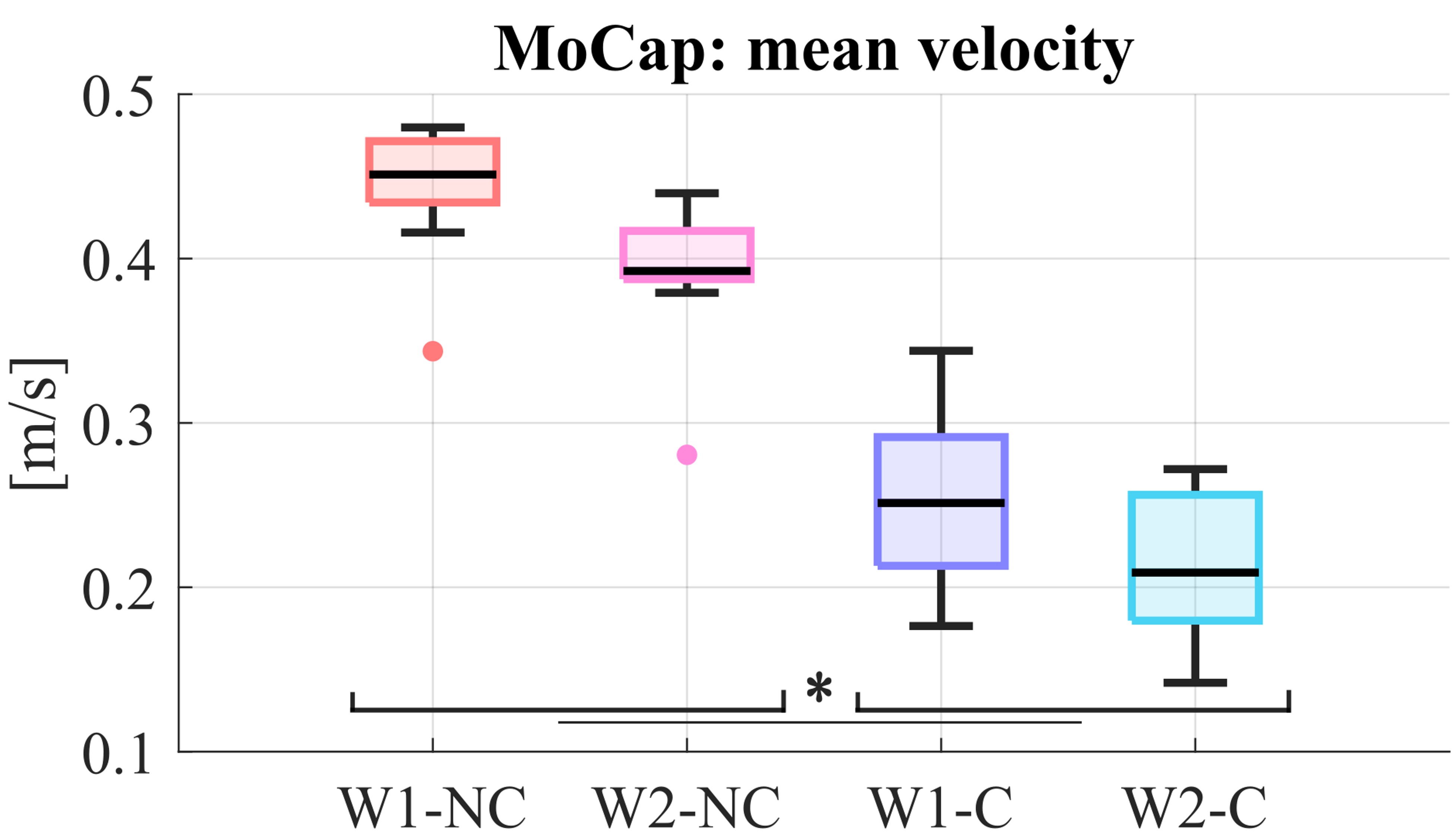}
    \caption{Mean velocity of the participants hand while manipulating the four different glasses, acquired through active infrared markers. The black lines represent the medians, the colored rectangles limit the 25\textsuperscript{th} and 75\textsuperscript{th} percentiles, and $*$ indicates a significant difference according to the Kruskal-Wallis test.}
    \label{fig:meanVel}
  \end{subfigure}
  \hfill
  \begin{subfigure}[b]{0.45\textwidth}
    \centering
    \includegraphics[width=\linewidth]{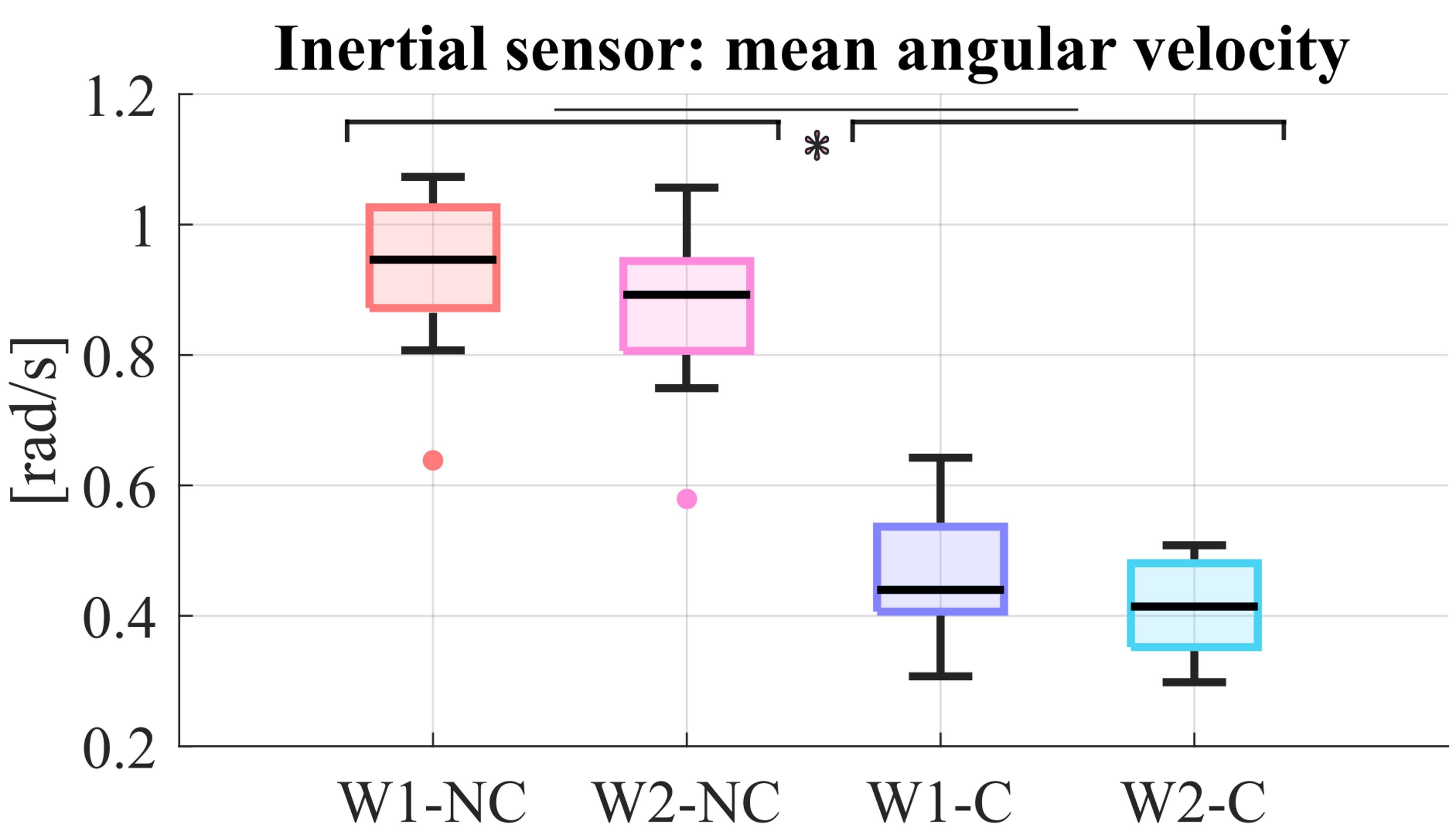}
    \caption{Mean angular velocity during the glasses transportation as recorded by the smartwatch on the participants right wrist. Same graphical conventions as in Figure \ref{fig:meanVel} above.}
    \label{fig:angVel}
  \end{subfigure}
  \caption{Kinematics parameters during the transport movements for the four different classes, retrieved for motion capture data (\ref{fig:meanVel}) and inertial sensor data (\ref{fig:angVel}). Figure (\ref{fig:evolution}) shows as a proof of concept how the mean velocity measured with the moCap stabilizes when considering a properly size pool of participants.}
  \label{fig:inertial_mocap}
\end{figure}
\indent An in-depth kinematics analysis can be carried out to study how the properties of the different objects affect the movement of the arm and condition how the transport action is completed. Our dataset allows not only to calculate different kinematic parameters, but also to compare the information which can be retrieved from the different sensors. In Figure \ref{fig:inertial_mocap} can be found an example, which compares kinematics parameters extracted from the synchronized MoCap and inertial sensors during the transportation of the four glasses. Figure \ref{fig:evolution} illustrates, as a proof of concept, how the hand mean velocity during the transport phases evolves depending on the number of participants considered. As the sample becomes larger, the measure stabilizes, and from the point when the dataset has 11 subjects, the velocity is almost completely constant. This analysis hints that the sample size of 15 participants (for a total of 1200 trials) is suitable for a reliable description of the proposed pick-and-place scenario; indeed, the acquired data have an internal coherency and well represents the diversity and the natural variance of human motion in this context.
In detail, in Figure \ref{fig:meanVel} are shown the median, the 25\textsuperscript{th} and the 75\textsuperscript{th} percentiles of the hand velocity calculated deriving the 3D position of one of the markers placed on the volunteer's hand (markers 1 to 4, see Figure \ref{fig:markers} for reference). According to the Kruskal-Wallis test for non-normal distributions, we found a significant difference in the velocity adopted for transporting the glasses between those filled with water (W1-C and W2-C) and those empty, with simply a weight difference (W1-NC and W2-NC). Even though a trend is visible, no significant difference was found concerning the weight. The same results emerge in Figure \ref{fig:angVel} as well, where the wrist's mean angular velocities are recorded with the inertial sensor in the smartwatch. Again, a significant difference in the magnitude of the angular velocity appears between the Careful and Not Careful transport motions.\\
\begin{figure}[t]
\centering
\includegraphics[width=\columnwidth]{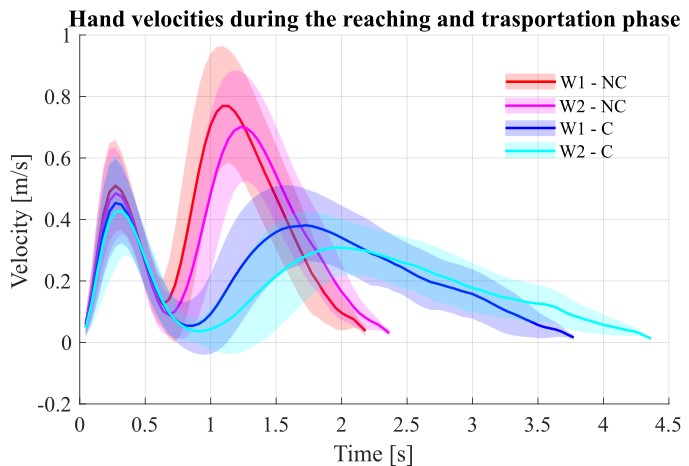}
\caption{Hand mean velocities and standard deviation, in transparency, associated with
the reaching (first peak) and transportation movements (second peak) of the four different glasses.}
\label{fig:v_profiles}
\end{figure}
\indent Finally, Figure \ref{fig:v_profiles} provides an insightful qualitative overview of the hand velocity profile acquired with the MoCap system during the reaching towards the cups (first peak) and the transport phase (second peak). To create such a global representation, the mean of the hand velocities of all the trials, separated into the four classes of motion, was computed for every time instant, together with its standard deviation. It can be noted how not only the maximum velocity decreases when transporting the full cups, but also the peak is anticipated, especially for the full, heavy container. Participants tended to be especially cautious in the final phase of the movement, gently leaning the glass so as not to spill the content.

\section{Code availability}\label{sec:codeavailability}
The Github repository\footnote{\url{https://github.com/lindalastrico/objectsManipulationDataset}} includes a number of MATLAB scripts allowing users to load and process the data. Further details on how to use the available functions are contained in each of them.
\begin{itemize}
\item \texttt{loadMocap} and \texttt{loadInertial} allow users to load and save, respectively, the motion capture and the inertial data coming from the smartwatch in an easy-to-use data structure in the form of a $15\times80$ cell array (number of subjects $\times$ number of trials). 
\item \texttt{loadTimestampsCameras} saves in a data structure the YARP timestamp for each one of the three cameras during the experiment.
\item \texttt{visualizeHandTrajectory} produces a 3D plot with the visible markers positioned on the hand for specified trial and subject combo.
\item \texttt{createVideo} creates a video from the images, specifying the desired participant, trial and camera (one of the two high resolution cameras or the robot camera). It also saves the YARP timestamps corresponding to the considered frames.
\item \texttt{animatedPlot} renders the video previously created with \texttt{createVideo} together with the trajectories of the markers on the hand during the specified trial and the three components of the acceleration recorded with the smartwatch. The 3D trajectories, the acceleration and the video are reproduced simultaneously; the number of the markers to visualize can be easily modified.
\end{itemize}

\section{Example of Usage}
\label{sec:exampleofuse}
In this section, we provide insights into the dataset usage. In particular, we compare the performances of the same classifier, a Long Short Term Memory Neural Network (LSTM-NN), applied to the data sources present in the dataset, i.e., MoCap, Robot Camera, and IMU. In this example, we focus on transportation motions because of the direct influence of the objects' physical characteristics given by the contact presence. After being segmented, the transportation phase of each trial was appropriately labeled according to the features of the cup involved; thus, we trained binary classifiers to discriminate its weight or the presence of carefulness in the motion. This study's objective is to provide insight on which sensing approach better fits the problem and if fusing information from different sensing modalities could help.

\subsection{Data Preprocessing}
Before using them to train and test the LSTM-NNs, it is necessary to preprocess the data. This preprocessing involves segmentation, feature extraction, data filtering, resampling, data normalization, and padding.

As shown in Figure \ref{fig:sequence}, each trial contains the whole pick-and-place action divided into reaching, transportation, and departing. At the beginning and the end of the transportation, the subject hand stops to grasp and release the object. This characteristic can be leveraged to isolate the transportation motion. We did it by computing the norm of the hand velocity, using the MoCap data, and applying a thresholding mechanism to identify the timestamps related to the start and end of the transportation. We used the resulting timestamps to segment the IMU and the robot camera data.

Given the different nature of the three sources of data selected, to adopt the same classification pipeline, it is necessary to extract consistent features. From the robot camera images, we computed the Optical Flow (OF) \citep{opticalFlow}, and we used it to extract the components of the motion velocity on the image plane. From these data, we then calculated the norm velocity, the angular velocity, the curvature, and the radius of curvature \citep{HFR}. We selected these four features since, in the past, they have been successfully adopted to characterize biological motions \citep{opticalFlow} and discriminate between careful and non-careful transportations \citep{HFR}. Instead, we selected six features both for the MoCap and the IMU. For the MoCap, we computed the hand triaxial linear acceleration and velocity. For the IMU, we used the raw data, i.e., linear acceleration and angular velocity.

We suggest filtering the data for noise reduction as part of the data preprocessing. In this case, we applied to every temporal sequence a first-order Butterworth filter with a threshold frequency equal to the original sampling rate of the sensor (i.e., \SI{71}{\hertz} for the IMU, \SI{100}{\hertz} for the MoCap, and \SI{22}{\hertz} for the camera). To simplify the comparison and easily combine information from the different sources, we resampled the IMU and MoCap data to match the camera sampling frequency. This choice also finds support in previous research suggesting that \SI{20}{\hertz} is an ideal sampling frequency for the perception of human daily activities \citep{frequencyResample}. To perform the resampling, we interpolated the data and used the camera timestamp to extrapolate the new data.

The resulting sequences are then scaled using min-max normalization to decrease the difference in scale between the features. For the IMU, we selected as maximum and minimum values the full-scale range of the sensors, i.e., $\pm$\,2g for accelerations and $\pm$\,\SI{8.73}{\radian/\second}  ($\pm$\,500 deg/s) for the angular velocities. Finally, since all the trials have a different temporal length, we padded the data with zeros to be able to use batch training. In particular, we used pre-padding since it is considered more robust to the noise introduced by the zeros \citep{padding}.

\subsection{Model Training and Validation}


As mentioned before, our experiment aimed to compare the results of different classifiers for distinguishing, separately, whether carefulness was adopted during the transportation and if the object involved was heavy or light. The evaluation focuses on four classifiers that differ from each other for the used data source. The study comprehends one classifier for each sensor (i.e., camera, IMU, and MoCap) and a classifier using both IMU and camera data. We explore the combination of IMU and camera data to determine if an autonomous robot could leverage it for more reliable perception. We chose a simple LSTM model followed by fully connected layers for each of the four models. The networks were implemented in Python using the sequential layers provided by Keras\footnote{\url{https://keras.io/}}.
The LSTM layer has 64 hidden units and an input shape equal to [$sequence\_length \times n\_features$], where the sequence length is fixed to 134 samples (maximum sequence length after resampling), and the number of features varies according to the data source (i.e., 6 for MoCap and IMU, 4 for the camera and 10 for IMU plus camera). The next layer is fully connected, with 32 neurons, and it is preceded and followed by dropout layers with a value of 0.5. The output layer is another fully connected one with two output neurons corresponding to the two classes: careful and not careful. Given the double output, we chose a softmax function for the activation and the categorical cross-entropy to evaluate the loss. An L1-L2 kernel regularization was added to the last layer to prevent the model from overfitting with $L1 = 0.001$ and $L2 = 0.001$ as parameter values. The chosen optimization algorithm was AdamOptimizer, with a learning rate of 0.0002 and batch size of 16.


For each of the four models, we carried out the training and testing phases by adopting the Cross-Validation with a Leave-One-Out approach to test the ability of the model to generalize over different participants. Therefore, to split the 1200 sequences ($15\;participants \times 80\;sequences$) into training, validation and test sets, we adopted the following procedure. One at a time, the data of each participant were used as a test set, and the remaining 14 were further divided, 80\% for the training and 20\% for the validation. The validation set has been picked randomly from the 14 volunteers. The models have been trained for 100 epochs using an early stopping on the validation loss with patience set to 5 epochs to avoid overfitting.

\subsection{Results}
\begin{figure}[t!]
\centering
\begin{subfigure}[b]{0.45\textwidth}
\includegraphics[width=1\textwidth]{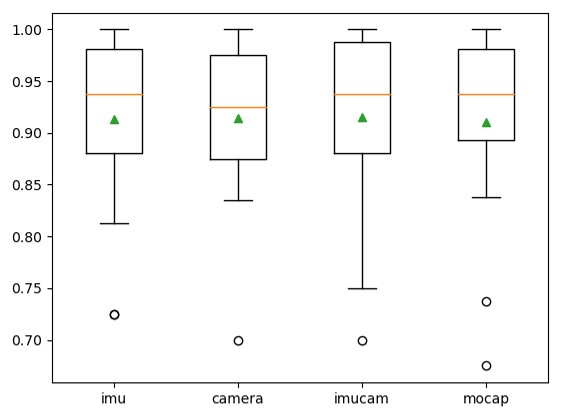}
\caption{Carefulness classification}
\label{fig:caref_results}
\end{subfigure}
\begin{subfigure}[b]{0.45\textwidth}
\includegraphics[width=1\textwidth]{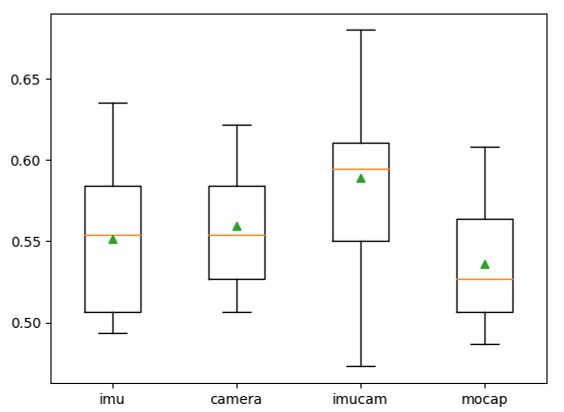}
\caption{Weight classification}
\label{fig:weight_results}
\end{subfigure}
\caption{The results of the four classifiers trained with different data sources in terms of accuracy expressed as boxplots for carefulness (\ref{fig:caref_results}) and weight (\ref{fig:weight_results}) discrimination. The yellow line represents the median, while the green arrow indicates the average.}
\label{fig:results}
\end{figure}


The classification results of the four models are reported in Figure \ref{fig:results} using boxplots showing the median, the average, and the distribution of the values for each data source. Regarding the carefulness, the overall performances of the models are comparable, i.e., 91.3\% for the IMU, 91.4\% for the camera, 91.6\% for the MoCap, and 91.1\% for IMU plus camera (see Figure \ref{fig:caref_results}). Therefore, considering this evidence, the 4 data sources are almost equivalent for classification purposes. On the other hand, if we exclude the outliers, the minimum values displayed in the chart appear to reflect some differences between the models. The lowest accuracy achieved while using camera and MoCap data is around 84\%, for the IMU reaches approximately 82\%, and for the combination of IMU and camera achieves 75\%. This result suggests that combining IMU and camera data may not be particularly effective for estimating motion carefulness. However, since all the single classifiers reached high performances, it is possible to imagine an autonomous system combining the results of different classifiers to obtain more stable and reliable inferences.

With the same approach, the accuracy of the weight classification is instead not as satisfying as the one achieved for carefulness. The mean accuracy values, as reported in Figure \ref{fig:weight_results}, are 55\% for the IMU, 56\% for the camera, 54\% for the MoCap, and 59\% for IMU plus camera. This result is particularly interesting, as the inference process for the weight is not as simple as expected. Our hypothesis is that the greatest challenge for the volunteers during the experiment was to safely handle the filled glasses and that the difference in weight between the objects did not have an impact as strong as the water filling on the kinematics. This observation is confirmed by Figures \ref{fig:inertial_mocap} and \ref{fig:v_profiles}, where the significant difference in the considered kinematics features emerged only between full and empty cups. Objects can have multiple concomitant features, which do not always have the same effect on kinematics as their interaction may lead to the attenuation of their effect as compared to when considered individually. Our dataset, by attentively balancing the combinations of water filling and weight along multiple directions, can be particularly useful to deepen the understanding of human motor strategies in this context.

\subsection{Discussion}
Based on the results presented in the previous section, it appears to be inconsequential which sensing modality we adopt, regardless of the object feature under study. However, we should make a few observations to provide a more comprehensive view of the matter.
Firstly, we should note that, in our approach, we processed RGB images with the optical flow to extract features that describe only human motion. We proceeded in this way to isolate human kinematics, but this procedure reduces the richness of information of the sensing modality. In fact, we would expect a classifier trained on raw images to achieve better accuracies since it can leverage the visual features of objects as well.
Secondly, the presented results have been achieved in a specific experimental setting designed to be as fair as possible for each sensing modality. Therefore, future studies could benefit more from specific sensing modalities depending on the constraints and characteristics of the application and human actions under consideration. For example, in a scenario where humans interact with crowded environments or use different motor actions, the frequent occlusions with other objects could impair visual sensing. On the other hand, wearable sensors such as IMUs are not affected by occlusions, but it is not always possible or convenient to sensorize humans.

\section{Conclusions}
This article provides a multimodal dataset of human object pick-and-place tasks under different experimental conditions. In particular, the dataset describes the effect of object weight and associated carefulness on human motions during a pick-and-place. The dataset is collected in a controlled environment with 15 subjects performing 80 different pick-and-place actions. The dataset contains multiple camera views, IMU, and MoCap data providing the possibility of integrating and comparing the diverse sensing modalities. The data collection was prompted by the need to create reliable models of human strategies in the context of object manipulation, which could be used by robots to understand the scene and facilitate interaction. In this article, as an example of usage, we propose a simple comparison of the performances of classifiers using different sensing modalities to distinguish between careful and non-careful transportations or between heavy and light cups.

The motions which constitute the dataset are also strictly controlled concerning their direction (left, right, up, down, adduction, receding) and the different phases of the actions can be easily split, distinguishing between the reach-to-grasp, the transport, and the departing motions (see schema in Figure \ref{fig:sequence}). This makes the dataset particularly suitable for studying human intention recognition; understanding what is going to be grasped, and where, can lead to significant improvements in the HRI experience, allowing robots to anticipate our goals and adapt accordingly. This kind of studies can also be conducted by comparing the different sensors available, so as to build a framework that can be adapted to different contexts: for instance, by relying more on inertial data in case of obstructions to the cameras.\\
Other applications to Robotics fall under the theme of motion generation. The presented dataset, however, is not aimed at classical robot learning from human demonstration; indeed, useful sensors for such use, such as grip and force sensors, are not included. Still, the acquired human velocity profiles have been used to build on top of standard robot trajectories (pick-and-place and collaborative handovers) by adding a communicative layer to the motions. Generative Adversarial Networks were trained on human examples to produce velocity profiles falling within the desired careful/not careful attitude and such time-series used to control the end-effector trajectories. Such an approach has been successfully deployed on humanoids and robot manipulators producing gestures adapted to cups' content and improving the interaction efficiency (see \cite{ICDL, Lastrico2023IROS} for reference).

This dataset represents a contribution to the study of how specific object characteristics influence human motion. In future works, it could be interesting to expand the dataset to consider different levels of carefulness and study which factors (such as context, value, fragility, potential danger, and so on) impact human motion going beyond the spillability of the water content.

\section{Declaration of competing interest}
The authors declare no competing interests.

\section{Acknowledgements} 
A.S. is supported by a Starting Grant from the European Research Council (ERC) under the European Union’s Horizon 2020 research and innovation program. G.A. No 804388, wHiSPER. 
This work is partially supported by the CHIST-ERA (2014-2020) project InDex, and received funding from the Italian Ministry of Education and Research (MIUR).
This research is partially supported by the Italian government under the National Recovery and Resilience Plan (NRRP), Mission 4, Component 2 Investment 1.5, funded by the European Union NextGenerationEU and awarded by the Italian Ministry of University and Research.

 \bibliographystyle{SageH} 
 \bibliography{cas-refs}






\end{document}